# Shape-programmable Magnetic Soft Membranes for Mechanically Active Microchannels

Direnc Akyildiz[1], Fatih Kocabas[1], Xiangyi Tan[1,2] and Yunus Alapan[1,3,4]

*Abstract*— **The capability to encode spatially distinct magnetization patterns within soft materials enables remote control over complex deformations. This characteristic is especially important for microfluidic platforms, where limited dynamic control of channel boundaries and laminar flow conditions usually restrict fluid transport and interactions. The present study introduces a shape-programmable magnetic soft membrane actuator as an active microchannel component that can dynamically modulate its shape under magnetic fields and therefore the microfluidic environment. The membrane is magnetically programmed using a template-based approach, which allows it to be controllably deformed in the form of a sinusoid under the influence of an external magnetic field. The membrane's integration into a microchannel converts a passive channel wall into a dynamically changeable interface, allowing active fluid manipulation and enhancing micromixing in laminar flow conditions. The proposed approach establishes a versatile platform for wirelessly controlled deformable interfaces in next-generation microfluidic and lab-on-chip systems.**

## I. Introduction

Magnetically actuated soft materials have gained substantial interest in a vast variety of fields, ranging from soft robotics to biomedical devices and adaptive engineering systems due to their ability to provide contactless actuation and capability to generate complex shape-morphing under external magnetic fields [1-4]. These materials, consisting of ferromagnetic microparticles embedded in a soft polymer matrix, enable remotely actuated, pre-programmed, and reversible deformations via spatially distributed magnetization profile, thus distributed magnetic forces and torques under external magnetic fields [5-11]. The capability to program complex shape-morphing has been widely explored in small scale soft robotics for multi-modal locomotion, targeted drug delivery and tissue manipulation [2, 3, 12,13].

Another potential area that could greatly benefit from such shape-morphing material systems is microfluidic and lab-on-a-chip platforms. Microfluidic technologies have been widely employed in drug discovery, fabrication of microparticles/emulsions, micromixing, cell/analyte capture from complex biofluids, and cell culture under physiological conditions [14-18]. Efficacy of microfluidic systems are constrained by fluid physics governed at microscale, where channel diameter or height span tens to hundreds of micrometers with intrinsically low Reynolds numbers ($Re = \rho V D_h/\mu$), leading to laminar flow that restricts fluid flow interactions. This limitation becomes especially critical for applications such as micromixing, separation of cells/analytes, and cell-based sensing platforms, where the performance is often affected by fluid-channel interactions. Overcoming this constraint usually requires complex channel geometries or application of external forces to create intricate flow patterns that enhance fluid interactions [15]. In conventional microfluidic platforms, forces are generally exerted through deformation of thin membrane systems integrated into channel walls and triggered via pneumatic pressure [19]. However, requirement of additional pneumatic channels increases complexity and limits robustness of pneumatically actuated membrane systems.

On the other hand, wirelessly activated magnetic soft membranes have been explored as diaphragm-based pumps and active valve components [20-22]. In addition, magnetically actuated soft membranes were further used for active mixing in microfluidic channels [23]. Magnetic soft membranes have been shown to generate periodic deflections created under external magnetic fields and improve the mixing performance. However, these systems employ simple and uniform deformations of the membrane structure, akin to pneumatically actuated valves and pumps. As a result, most of the existing magnetic micromixers and membrane actuators are limited to global vibration or uniform deformation mechanisms, thus pre-programmed shape transformations cannot be obtained within the channel structure.

In this study, we introduce shape-programmable magnetic soft membrane actuators for mechanically active microchannels. Magnetic soft patches integrated within continuous PDMS (Polydimethylsiloxane) membranes are fabricated via micromolding, magnetically programmed within an impulse magnetizer, and bonded to microchannels, details of which are described in Section II. Effect of magnetic particle ratio within magnetic patches to deformation performance, sinusoidal shape-morphing within closed microchannels, and application of shape-morphing magnetic soft membranes in microfluidic micromixing are provided in Section III, along with concluding remarks in Section IV. Magnetic soft membrane strategy proposed here offers real-time reconfigurable microchannel walls with complex shape-morphing capabilities, with wide ranging applications in mixing, pumping, and biophysical stimulation of cells.

[1]D. Akyildiz, F. Kocabas, X. Tan, and Y. Alapan are with the Bio-integrated Robotics Lab at Mechanical Engineering Department, University of Wisconsin -Madison, Madison, WI 53706 USA (alapan@wisc.edu).

[2]X. Tan is with the Department of Electronic and Electrical Engineering, University College London, London, WC1E 7JE UK.

[3]Y. Alapan is with the Biomedical Engineering Department, University of Wisconsin - Madison, Madison, WI 53706 USA.

[4]Y. Alapan is with the Carbone Cancer Center, University of Wisconsin - Madison, Madison, WI 53706 USA.

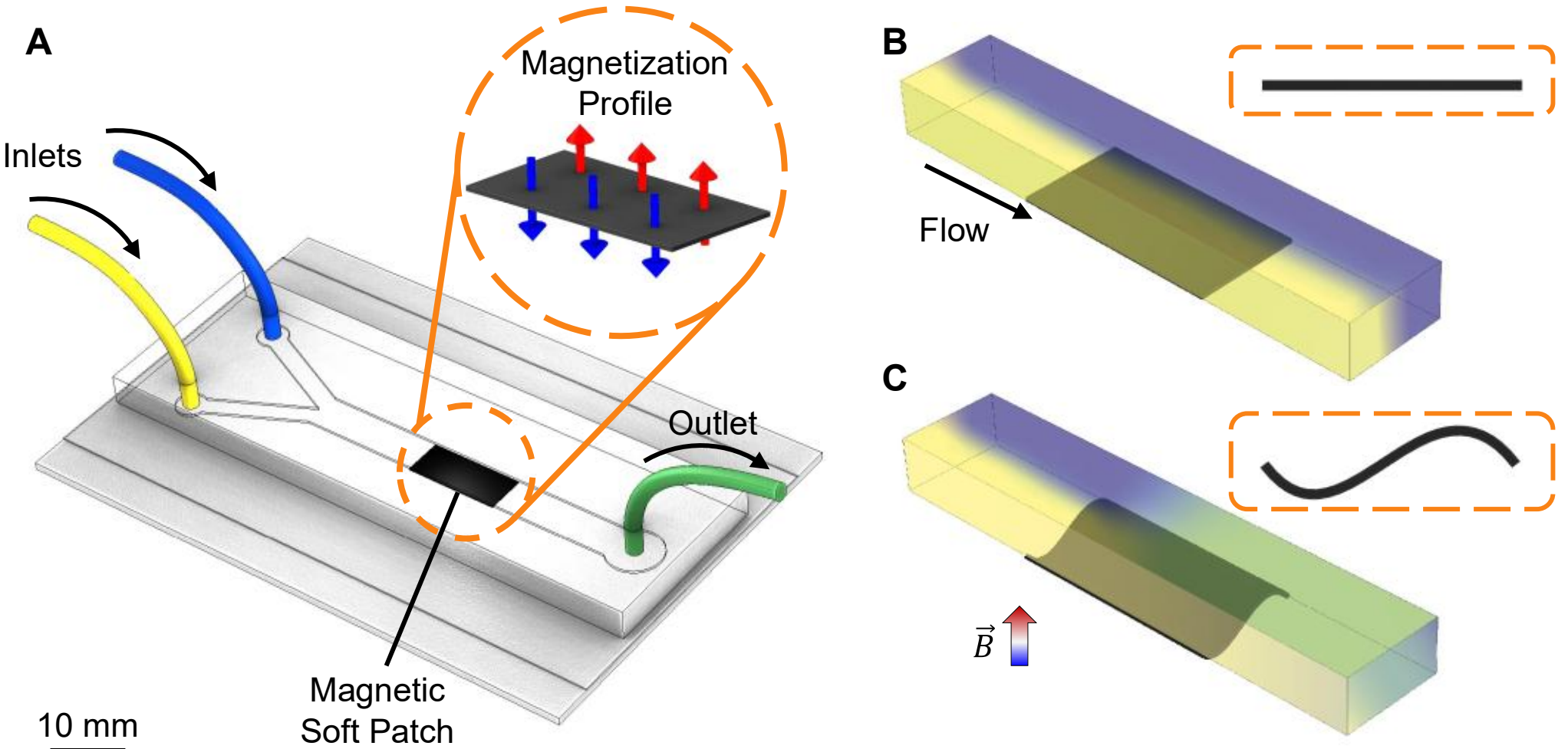

Figure 1. Shape-programmable magnetic soft membrane–integrated microchannel and its magnetically induced actuation mechanism. **A)** Schematic illustration of the complete microfluidic device consisting of two inlets and one outlet, with the shape-programmable magnetic soft membrane integrated into the microchannel. The inset shows the programmed magnetization profile of the magnetic soft patch. **B)** Flow pattern inside the microchannel with the magnetic membrane in the non-actuated state, showing no observable mixing before and after the membrane region. The inset shows the cross-sectional profile of the non-actuated membrane, where no deformation is present. **C)** Flow pattern inside the microchannel after magnetic actuation of the membrane, where mixing is observed over and downstream of the membrane region. The inset shows the cross-sectional profile of the magnetically actuated membrane exhibiting sinusoidal deformation.

## II. Materials and Methods

### A. Concept

This work presents a continuous magnetic soft membrane designed as an active interface for adaptive microfluidic systems. Under externally applied magnetic fields, the proposed membrane's compliant soft structure and spatially programmable magnetic characteristics allow for controlled deformation, thereby enabling the membrane to function as a soft actuator that uses magnetic fields to produce consistent mechanical movements for fluid flow control in microscale environments (Fig. 1).

### B. Fabrication and Programming of Magnetic Soft Membrane

The shape-programmable soft magnetic membrane was fabricated from PDMS using a two-step process (Fig. 2A). First, a PDMS - Sylgard 184 mixture (10:1 base to curing agent ratio) poured onto a 50.8 mm × 76.2 mm glass slide. To control the thickness of the membrane, strips of adhesive tape with a thickness of 200 µm were placed along two opposite edges of the glass slide, acting as spacers and guidelines. The PDMS mixture across the substrate using a razor blade, which was guided along the edges of the tape. The sample was cured at 110 °C for 10 minutes on a hot plate. After curing, the tape spacers were removed carefully, and the membrane was allowed to cool to room temperature. A rectangular area of 10 mm x 6 mm was then removed from the center of the PDMS membrane using a razor blade. To prepare the soft magnetic composite, NdFeB was mixed with 10:1 PDMS mixture in the weight ratio (wt) of 75%, spread into the excised section to create a continuous magnetic soft membrane, and then cured at 110 °C for 10 minutes.

Magnetic shape programming of the magnetic soft membrane is achieved by using an impulse magnetizer (ASC Scientific, Model IM-10) operating at 1.2 T (Fig. 2B). Prior to magnetization, the membrane is mechanically constrained by folding 180° along the midline parallel to the long edge, bringing opposing membrane surfaces into contact. This folding configuration enabled spatial encoding of magnetic orientation during impulse magnetization. The folded membrane was then positioned vertically inside the magnetizer and exposed to a 1.2 T uniform magnetic field. When external magnetic fields were applied to the system, the opposing magnetic domains within the same continuous patch created distributed magnetic forces and torques that caused the membrane to experience alternating deformations.

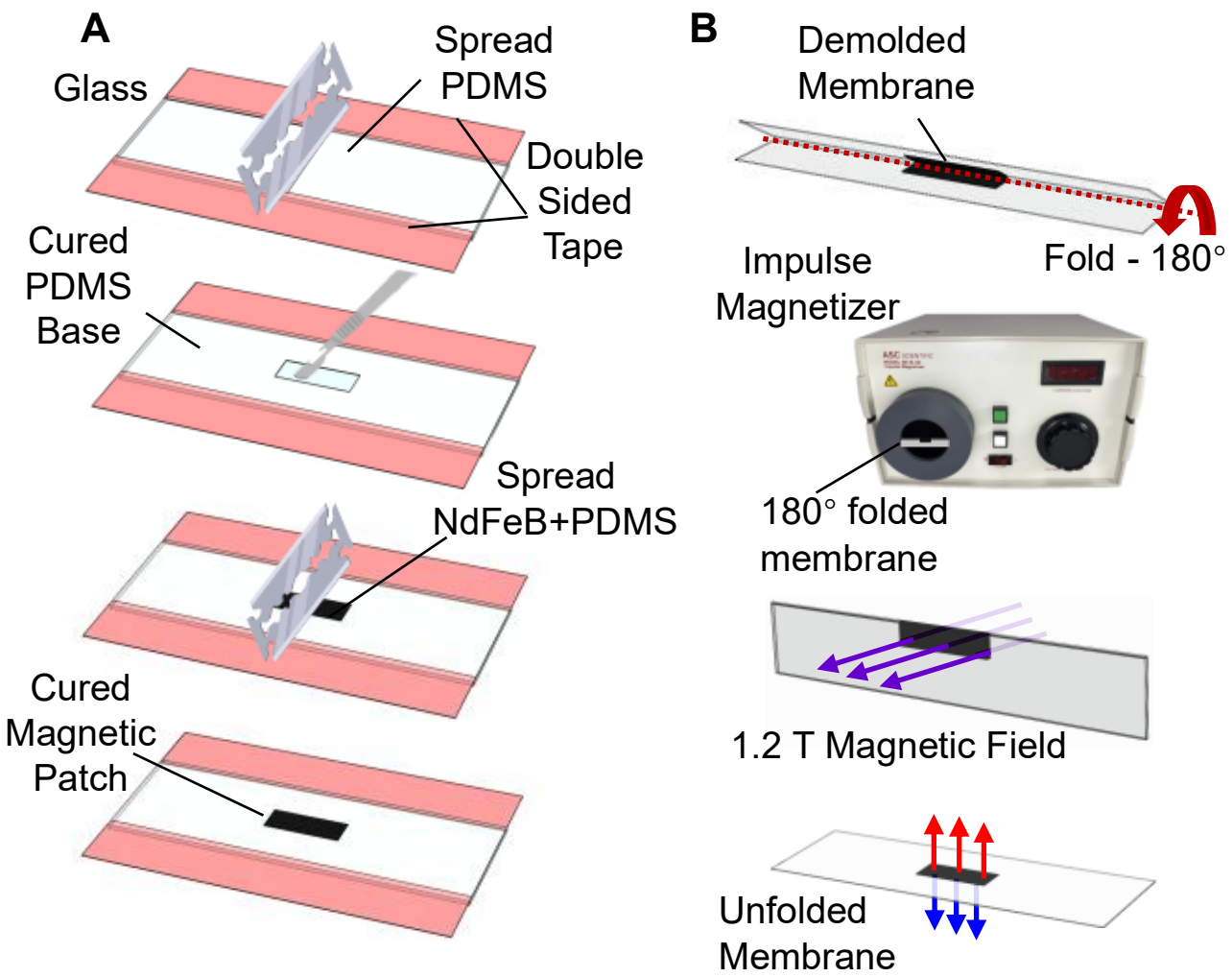

Figure 2. Fabrication and programming of the magnetic soft membranes. **A)** Fabrication of the membrane and magnetic patch. The PDMS layer is uniformly spread on a glass slide using 200 µm-thick spacer tapes as guides and then cured. A defined region at the center is then cut and excised, filled with an NdFeB–PDMS composite, and subsequently cured. **B)** Magnetic programming of the membrane. The magnetic soft membrane is folded 180° along the midline parallel to its long edge and magnetized using an impulse magnetizer, purple lines indicate magnetization direction. After unfolding, a 180° opposite magnetization profile is obtained, represented with red and blue arrows.

## C. *Fabrication and Assembly of PDMS Microchannel*

The PDMS microchannels were fabricated using a laser-cut double-sided adhesive tape mold defining a Y-shaped channel geometry, with inlet lengths and widths of 17 mm and 2.5 mm, respectively, a main channel width of 5 mm, and a channel length of 50 mm. The channel height (300 µm) was established through the attachment of two stacked layers of 3-mil double-sided adhesive tape (3M, USA) onto a glass slide, which served as spacers. A 10:1 PDMS mixture (Sylgard 184) was poured over the prepared mold and cured on a hot plate at 160 °C for 30 min. The PDMS layer was peeled off after the curing process, and the microchannels were formed through the removal of the patterned areas (Fig. 3A).

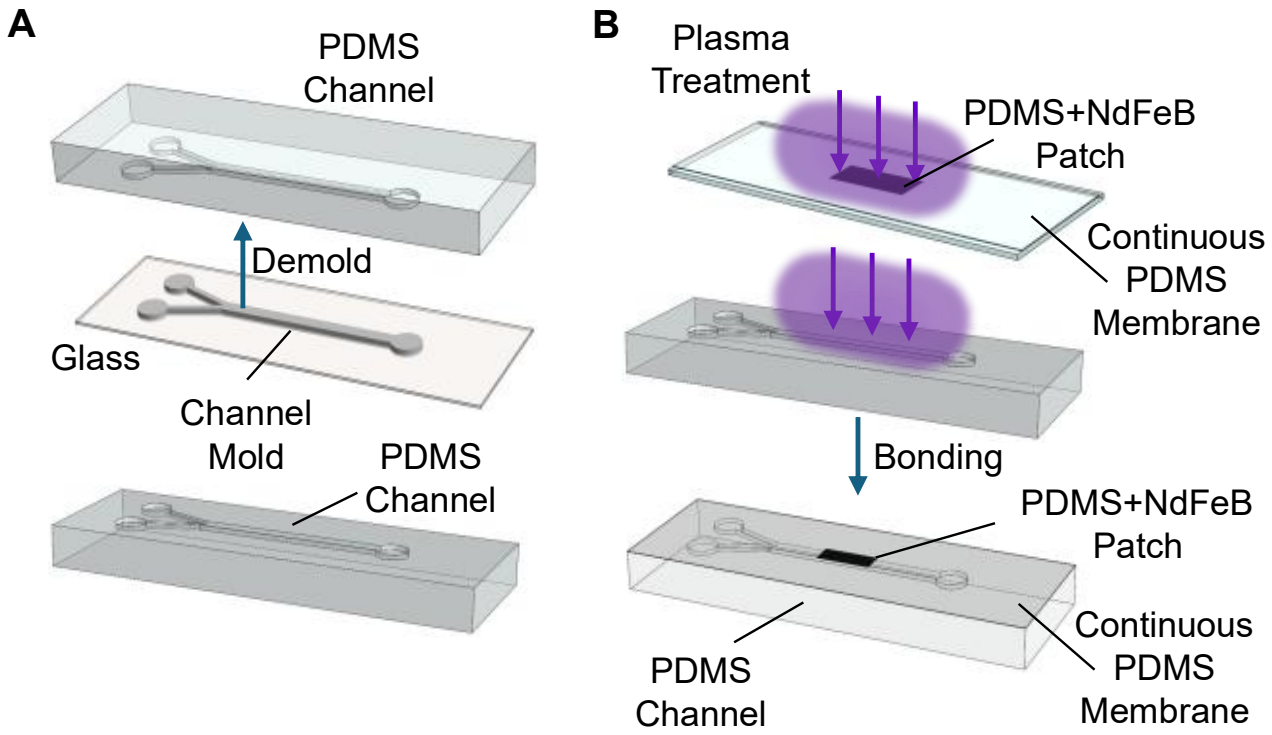

Figure 3. PDMS microchannel fabrication and device assembly. **A)** Fabrication of the microchannel. PDMS mixture is poured onto the channel mold on a glass slide, cured, and demolded to obtain the PDMS channel. **B)** Assembly of the microchannel. The surfaces of the PDMS channel and magnetic patch are plasma treated and bonded together to assemble the microchannel.

For microchannel assembly, both the PDMS channel and the membrane were cleaned with ethanol and bonded following oxygen plasma treatment (Fig. 3B).

## D. *Magnetic Actuation and Deformation Characterization*

Magnetic actuation–induced deformation of the soft magnetic membranes was characterized using ten microchannel–membrane assemblies (n = 5 for 50 wt% NdFeB–PDMS and n = 5 for 75 wt%). All membranes were magnetized using the programming method shown in Fig. 2B but without folding, resulting in a uniform magnetization orientation along the membrane. Consequently, the membranes exhibited single convex-concave (curved) deformation when subjected to external magnetic fields.

Microchannel cross sections were positioned vertically and imaged using a stereomicroscope (ZEISS Stemi 305, 2x front optics objective) equipped with a digital microscope camera (Swiftcam 25 MP) operated via Swift Imaging software. Static magnetic fields were generated by positioning 2 x N35 neodymium cylinder magnet (25.4 mm in diameter and 6.35 mm in height, Radial Magnets, Inc.) at predefined distances from the membrane to produce magnetic field strengths of 10 mT, 30 mT, 70 mT, and 150 mT, which were measured and set using a magnetometer (PCE-MFM 3500, PCE Instruments).

Magnet orientation was reversed to create push and pull actuation modes for each field level, which led to concave and convex membrane deformation, respectively.

Sinusoidal deformation of the shape-programmed membranes, obtained through the 180° folding-assisted magnetization procedure shown in Fig. 2B, was characterized using an optical profilometer (Zygo New view 9000). In accordance with the deformation performance of the uniformly magnetized membranes, only membranes made with 75 wt% NdFeB-PDMS were subjected to measurements. Surface measurements were obtained by scanning a localized section of the magnetic patch in the direction of the sinusoidal deformation that occurs along the channel width under magnetic activation. During measurement, the same permanent magnet configuration used in previous characterization was positioned beneath the microchannel assembly to generate an approximately 120 mT magnetic field. The optical profilometer was used to collect surface height data, which were then analyzed using ZygoMx software. After that, the generated .datx files were loaded into MATLAB (MathWorks, USA), where positional height information was taken out for additional analysis. For three separate samples, MATLAB was used to create three-dimensional surface reconstructions and deformation profiles, and the resulting surface measurement data were smoothed in OriginPro (OriginLab Corporation, USA) using a Savitzky–Golay filter while preserving the overall surface profile.

## E. *Experimental Setup*

An experimental configuration was implemented to enable controlled magnetic actuation of the membrane and subsequent observation of fluid behavior within the microchannel. To induce dynamic actuation of the programmed magnetic membrane, a rotating magnetic field was utilized to dynamically change the direction of the applied magnetic field. An actuation platform consisting of an N52 neodymium cylinder magnet (25.4 mm in diameter and 25.4 mm in height, KJ magnetics) mounted on a stepper motor using a 3D printed adapter was used for this purpose, allowing the magnet to rotate together with the motor shaft. The motor was controlled by an Arduino Uno microcontroller through a digital stepper driver (DM542T), powered by an external power supply, which served as the power interface between the controller and the motor. Control signals generated by the Arduino enabled periodic rotation of the magnet at a user-defined actuation frequency. The microchannel–membrane assembly was then fixed above the actuation platform.

Two fluids consisting of deionized water mixed with a water-based blue and yellow food dye (Herbeklab) at different concentrations were injected into microchannel using a syringe pump (InfusionONE 300 series, New Era Pump Systems Inc.) through separate inlet tubing while continuous discharge was maintained through outlet tubing. Both fluids contained Tween 20 at the same concentration (0.5% v/v) to reduce surface tension, while blue and yellow dyes were used at concentrations of 4.93 wt% and 5.35 wt% to provide optical contrast for visualization of mixing. The flow rate was set to 4.55 mL/h, corresponding to a Reynolds number ($Re = \rho V D_h / \mu$) of 0.5, ensuring laminar flow conditions. A camera (iPhone 16 Pro, Apple Inc.) positioned above the microchannel was used to record mixing of infused dyes within the microchannel.

*F. Quantitative Image Analysis of Membrane Deformation and Mixing Performance*

At each field strength (10, 30, 70, and 150 mT), deformation of uniformly magnetized membranes was captured and subsequently used for curvature analysis (Fig. 4A). Curvatures were quantified using ImageJ Fiji (National Institutes of Health) by tracking three nodes on the membrane: two located at the edges of the channel cross section and one at the point of maximum deformation at the center of the patch. The coordinates of these three points were denoted as $P_1=(x_1,y_1)$ and $P_3=(x_3,y_3)$ for the two edge nodes, and $P_2=(x_2,y_2)$ for the point of maximum deformation. These three points define a three-point circle whose radius (R) was calculated, and the curvature was defined as the reciprocal of the radius, ($\kappa = 1/R$) given in (1). For each image, an individual pixel-to-length calibration was performed to convert pixel measurements into physical dimensions.

$$\kappa = \frac{2\left|(x_2-x_1)(y_3-y_1)-(y_2-y_1)(x_3-x_1)\right|}{d_{23}\cdot d_{13}\cdot d_{12}} \quad (1)$$

$$\text{where } d_{ij}=\sqrt{(x_i-x_j)^2+(y_i-y_j)^2}$$

Visual characterization of the mixing process was made possible by the green coloration created by the combination of the blue and yellow dyes as the fluids mixed downstream of the magnetic membrane. This color-based mixing activity was quantified by analyzing the collected images with ImageJ Fiji. Images were transformed from RGB to HSV color space, where the hue channel represents the primary color information independent of brightness, reducing the impact of illumination fluctuations. The green pixel intensity was employed as a quantitative measure of mixing performance since green coloration is only observed when the two fluids are physically mixed. To determine the green pixel intensity inside a specific region of interest (ROI), the hue channel was divided into blue, yellow, and green regions and green pixel intensity, defined as the percentage of green pixels in the ROI, was extracted. Locally Estimated Scatterplot Smoothing (LOESS) function then was applied to the data to reduce oscillations caused by the 2 Hz operating frequency while preserving the overall trend of the intensity variation. ROI was defined near the outlet of the magnetic patch. This location was selected because the fluid passing through the magnetic patch fully occupies the microchannel at this position, ensuring that the ROI captures the mixed flow downstream of the actuated region.

## III. Results

*A. Deformation Response of Uniformly Magnetized Magnetic Soft Membranes*

We first characterized the deformation behavior of magnetic soft membranes depending on their magnetic particle weight ratio. Magnetic soft membranes fabricated with 50 wt% and 75 wt% NdFeB microparticle ratios were uniformly magnetized and exposed to varying levels of external magnetic fields and gradients, resulting in deformation of membranes (Fig. 4A). Under opposing magnetic actuation directions (push and pull), reversible convex and concave deformation modes were seen. Membrane deformation gradually increased in both actuation directions for both particle ratios as the applied magnetic field strength increased from 10 mT to 150 mT, exhibiting a consistent field-dependent response (Fig. 4A, Supplementary Video). The quantitative image analysis results support this observation, showing that the membrane curvature increases monotonically in both directions as the strength of the applied magnetic field increases (Fig. 4B).

The 75 wt% samples show larger values in curvature compared to the 50 wt% samples under same magnetic fields in both push and pull modes (Fig. 4A, B), demonstrating the enhanced performance of magnetic soft membranes with greater magnetic particle ratios. Based on this characterization, membranes fabricated with 75 wt% NdFeB–PDMS were selected for subsequent sinusoidal shape-programming experiments, including microfluidic mixing analysis.

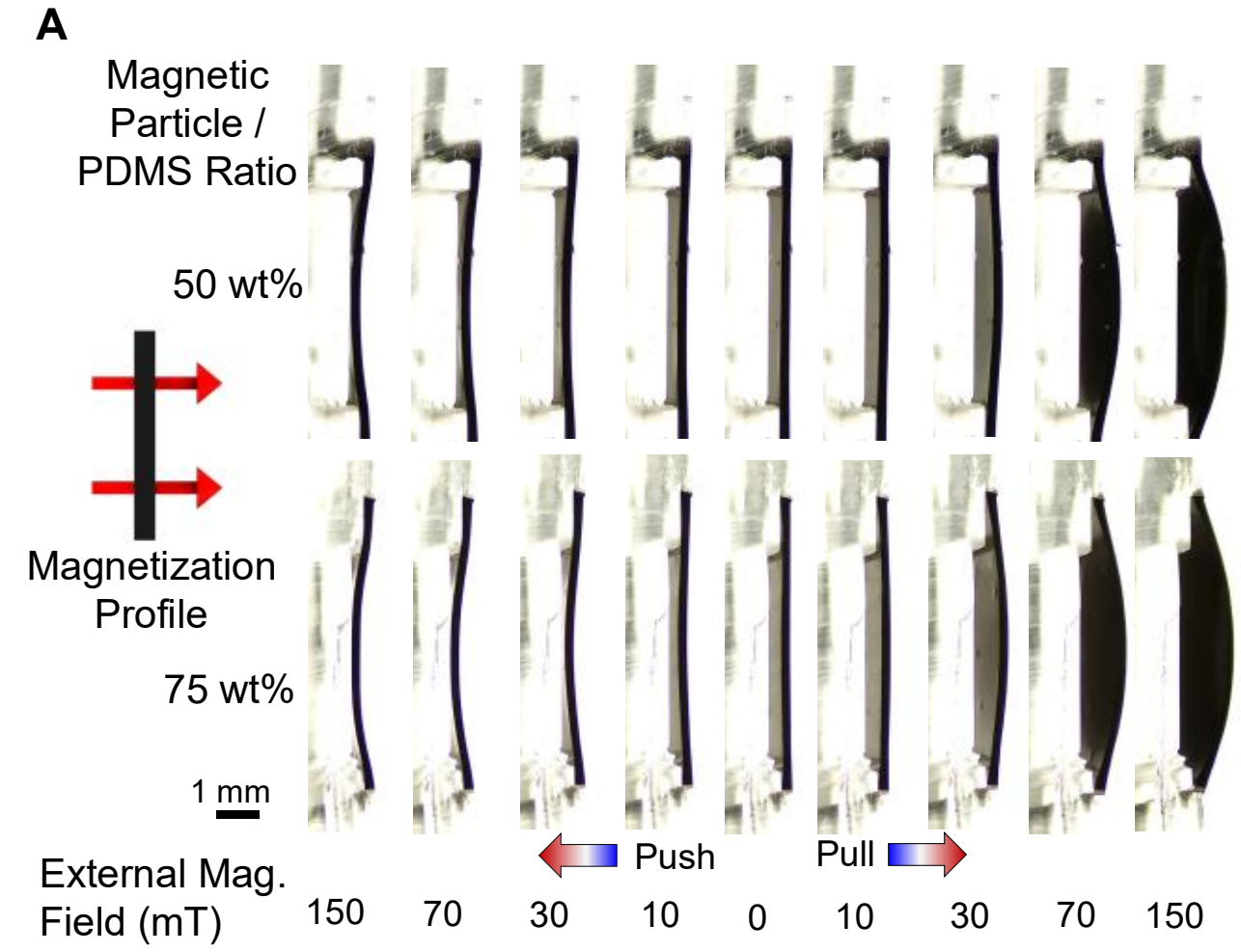

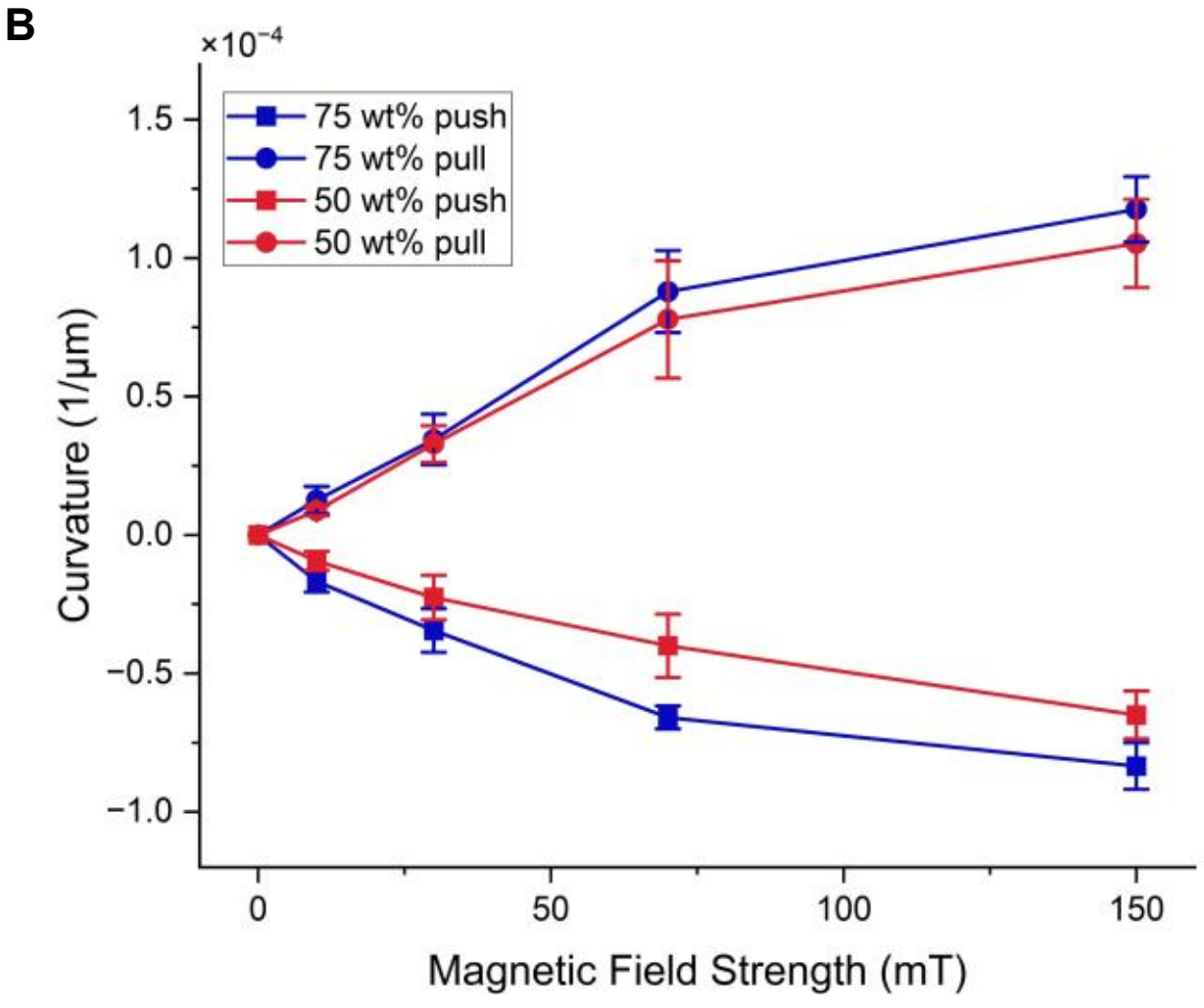

Figure 4. Characterization of uniformly magnetized patch deformation under magnetic actuation. **A)** Deformation of uniformly magnetized patch under magnetic fields. The upper row shows the NdFeB–PDMS (50 wt%) patch, and the lower row shows the NdFeB–PDMS (75 wt%) patch. Deformations are presented for magnet pushing, neutral/away, and pulling configurations at magnetic field strengths of 150, 70, 30, and 10 mT. **B)** Comparison of deformation curvature for NdFeB–PDMS patches with different particle weight ratios under the pushing and pulling conditions shown in **A**. Each data point represents the mean curvature obtained from five samples, with error bars indicating the standard error.

## B. Sinusoidal Deformation of Shape-programmed Magnetic Soft Membranes

Next, three-dimensional surface measurements acquired using optical profilometry (Fig. 5A) confirm the formation of sinusoidal deformation along the membrane width. A representative three-dimensional surface reconstruction obtained from one of the samples is presented in Fig. 5B. The deformation pattern persists throughout the scanned area, indicating a proper and continuous shape programming of the magnetic patch.

To understand reproducibility, three different samples were characterized. All samples showed similar sinusoidal deformation trends with comparable wavelengths and overall shapes (Fig. 5C). Quantitative analysis yielded an average absolute peak displacement of $146 \pm 22.48$ µm and a half wavelength of $1.94 \pm 0.38$ mm across the tested samples. This indicates consistent deformation behavior among the fabricated membranes. Small differences in displacement amplitude can be attributed to tolerances in fabrication and magnetic programming of soft membranes.

## C. Magnetically Induced Mixing

To demonstrate the applicability of the proposed magnetic soft membrane as an active microfluidic component, magnetically induced mixing was investigated as a representative application (Supplementary Video). Under the rotating magnetic field, the membrane showed periodic changes in its sinusoidal shape which caused the wave peaks and wave troughs to move up and down between their maximum and minimum points.

Mixing behavior induced by sinusoidal shape-morphing of the membrane was investigated through controlled experiments using the experimental set-up shown in Fig. 6A. Blue and yellow dye solutions were introduced from the two inlets of the microchannel, and an external magnetic field was applied by rotating a permanent magnet beneath the magnetic patch using a stepper motor at 2 Hz, generating a maximum field strength of 120 mT (Fig. 6A, B). The progression of magnetically induced mixing in the microchannel is shown in Fig. 6C, D.

Continuous magnetic actuation induces alternating deformation of the membrane, promoting mixing both within the magnetic patch region and downstream of it along the flow direction. This process results in the gradual development and growth of the green colored region. The progression of mixing along the direction flow is visually demonstrated by the growing dominance of green coloration between t = 5 s and t = 16 s (Fig. 6C, Supplementary Video). This observation is supported by the corresponding quantitative analysis shown in Fig. 6D, where the green pixel intensity within the ROI increases rapidly during actuation, rising from 6.0% to 80.0% over 13 seconds and reaching 93.1% by the end of actuation at 16 seconds.

# IV. Conclusion

In this paper, we present a shape-programmable magnetic soft membrane capable of acting as an active interface in microfluidic systems. The proposed membrane's spatially distributed magnetization profile allows it to create periodic sinusoidal deformation under the applied external magnetic fields. This dynamic shape-morphing within closed microchannels enables effective manipulation of the microfluidic flow environment through transforming the static channel border into a dynamic surface, offering a more versatile and multipurpose microfluidic platform. Magnetic actuation characterizations showed that membrane deformation changes in a controlled and reversible manner depending on the applied magnetic field strength and

A

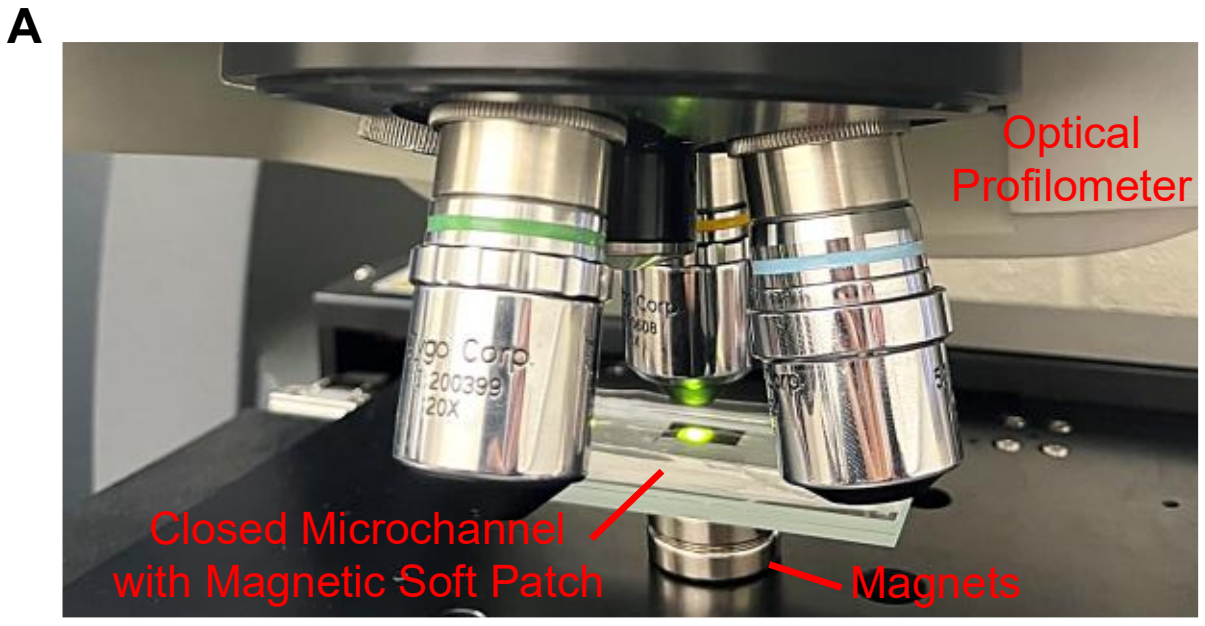

B

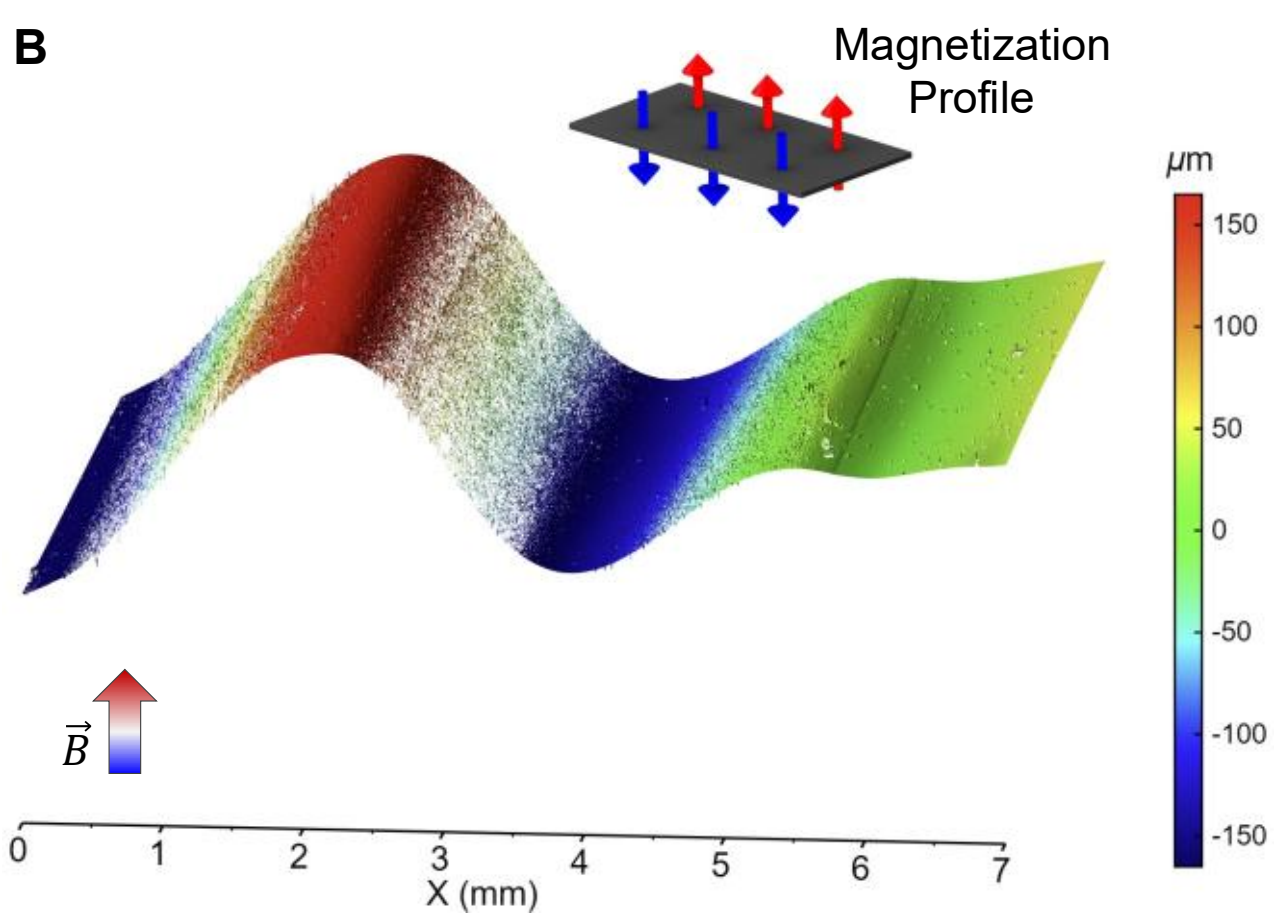

C

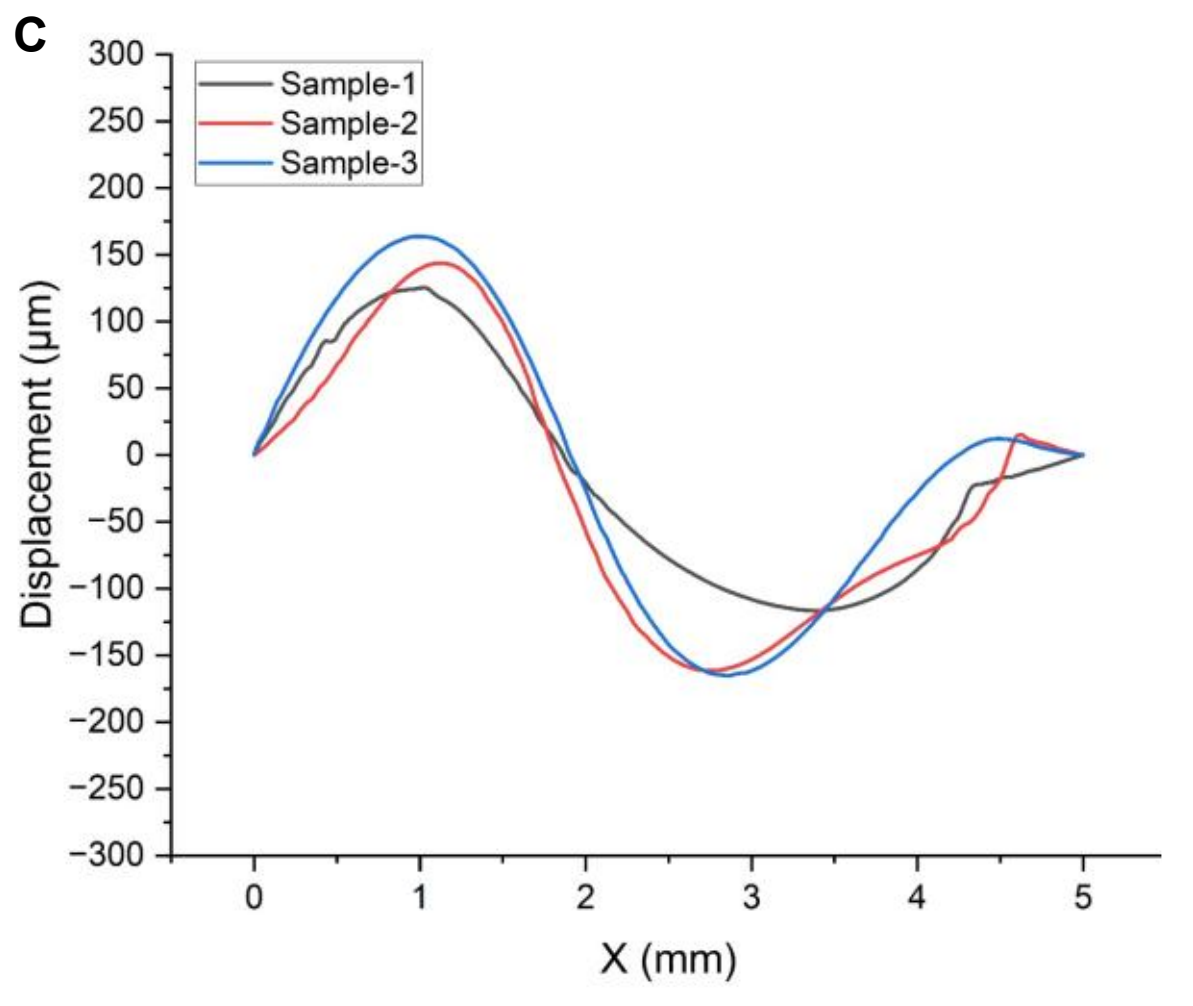

Figure 5. Characterization of magnetically induced sinusoidal deformation. **A)** Laser profilometer setup used for sinusoidal deformation measurements. The closed microchannel with magnetic soft patch is placed beneath the objective and actuated using external magnets during profilometry. **B)** Three-dimensional deformation profile of an NdFeB–PDMS (75 wt%) patch magnetized with 180° folding and actuated under an applied magnetic field of 120 mT, obtained from optical profilometry measurements. **C)** Optical profilometer measurements of the sinusoidal deformation for three NdFeB–PDMS (75 wt%) patch samples. The X-axis corresponds to the scan direction along the short axis of the patch during optical profilometry.

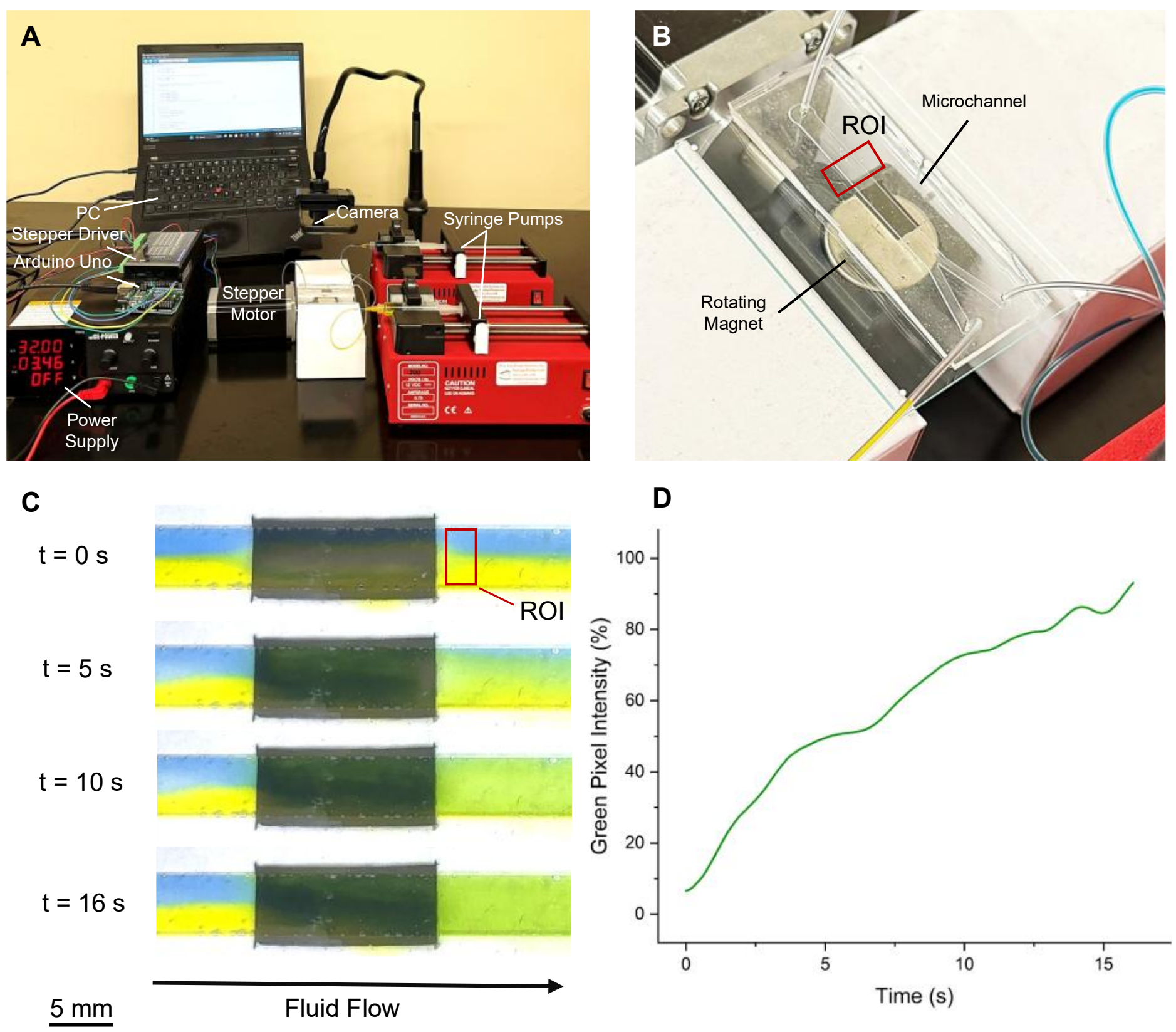

Figure 6. Micromixing experiment. **A)** Experimental setup for micromixing. **B)** Detailed view of the microfluidic setup and rotating permanent magnetic for external field generation. **C)** Captured microchannel images at t = 0 s, 5 s, 10 s, and 16 s after the magnetic soft membrane is activated under an applied magnetic field of 120 mT, initiated at t =0 s.; the ROI at the end of the magnetic patch near the outlet is labeled for subsequent image processing. **D)** Green pixel intensity within the ROI versus actuation time of the microchannel operating at 2 Hz under an applied magnetic field of 120 mT, corresponding to the region shown in **C**.

direction. These actively deformable surfaces can allow for dynamic control of localized flow regulation, cell/exosome transport and isolation in liquid biopsy applications, and dynamically adjustable biophysical environments in microphysiological and organ-on-a-chip systems without requiring complex channel geometry or mechanical components that are permanently connected. Future work may employ heat-assisted magnetic programming methods that enable high-resolution encoding to achieve more complex and discrete magnetization profiles in magnetic soft membranes [9-11]. The integration of these capabilities with closed-loop sensing systems may enable more adaptive and programmable microfluidic platforms.